\documentclass[letterpaper, 10 pt, conference]{ieeeconf}  

\IEEEoverridecommandlockouts                              

\usepackage[utf8]{inputenc} 
\usepackage[T1]{fontenc}    
\usepackage{hyperref}       
\usepackage{url}            
\usepackage{booktabs}       
\usepackage{amsfonts}       
\usepackage{nicefrac}       
\usepackage{microtype}      
\usepackage{xcolor}         

\usepackage{pgfplots}
\usepackage{comment}
\usepackage{amsmath}
\usepackage{hyperref}
\usepackage[caption=false,font=footnotesize]{subfig}
\usepackage{gensymb}
\usepackage{booktabs}
\usepackage{multirow}
\usepackage{siunitx}

\newdimen\figrasterwd
\figrasterwd\columnwidth

\title{\LARGE \bf
Regressing Transformers for Data-efficient Visual Place Recognition}

\author{Mar\'ia Leyva-Vallina$^{1}$, Nicola Strisciuglio$^{2}$, and Nicolai Petkov$^{1}$
\thanks{$^{1}$Mar\'ia Leyva-Vallina and Nicolai Petkov are with the Bernoulli Institute of the University of Groningen, the Netherlands
        {\tt\small m.leyva.vallina@rug.nl}}%
\thanks{$^{2}$Nicola Strisciuglio is with the University of Twente, the Netherlands
        {\tt\small n.strisciuglio@utwente.nl}}%
}

\begin{document}

\maketitle
\thispagestyle{empty}
\pagestyle{empty}

\begin{abstract}
Visual place recognition is a critical task in computer vision, especially for localization and navigation systems. Existing methods often rely on contrastive learning: image descriptors are trained to have small distance for similar images and larger distance for dissimilar ones in a latent space. However, this approach struggles to ensure accurate distance-based image similarity representation, particularly when training with binary pairwise labels, and complex re-ranking strategies are required. This work introduces a fresh perspective by framing place recognition as a regression problem, using camera field-of-view overlap as similarity ground truth for learning. By optimizing image descriptors to align directly with graded similarity labels, this approach enhances ranking capabilities without expensive re-ranking, offering data-efficient training and strong generalization across several benchmark datasets.
\end{abstract}
\section{Introduction}
Visual place recognition (VPR) is the task of identifying a previously visited location by analyzing visual information. It is an important component of visual navigation~\cite{Milford2012,Lowry2016} for autonomous driving and robotics~\cite{Doan2019}.
The task of visual place recognition is formulated as an image retrieval problem. Given a database of map images and a query image, the objective is to retrieve the most similar images to the query image from the database. The query image is recognized based on a distance metric between its descriptor and those of the database images~\cite{benchmarkingir3DV2020}. A key challenge of VPR is the ability to match images from different viewpoints and under varying appearance conditions. Thus, robust image descriptors have to be computed that take into account such issues~\cite{Zaffar2021}.
\begin{figure*}[t]
    \centering
    \vspace{2mm}
    \includegraphics[width=\textwidth]{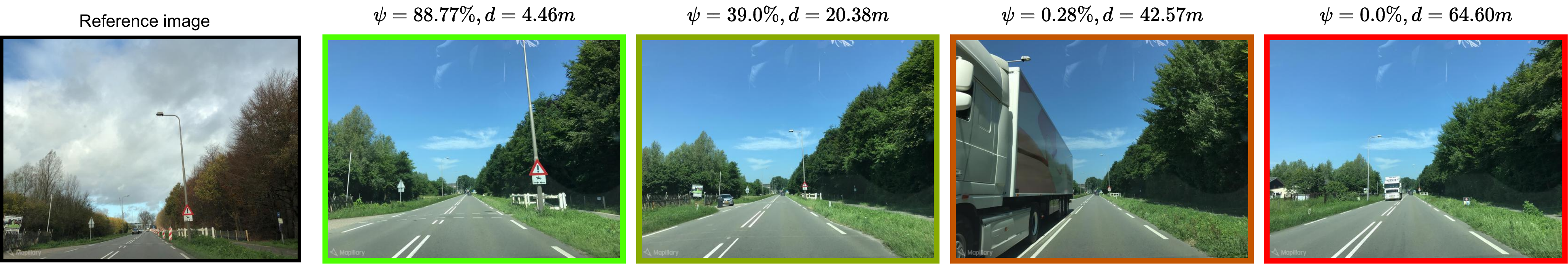}
    \caption{A reference image (leftmost) and four match images, taken at different distances. The larger the distance w.r.t the reference image, the lower the annotated similarity ground truth $\psi$, and the smaller the amount of shared visual features.}
    \label{fig:continuous_sim}
\end{figure*}

Top-performing methods were mostly based on convolutional backbones, trained with image pairs~\cite{radenovic2018fine} or triplets~\cite{Arandjelovic2017,lopez2019place}. In recent works, transformers replaced the convolutional backbones~\cite{wang2022transvpr,BertonCosplace,zhu2023r2former}. Training images in benchmark datasets are labelled so that they either depict the same place or not, from different points of view (distance and orientation of the camera) or at different times of the day or year. In this way, a certain level of robustness is expected to be embedded in the descriptors. However, the binary labels may introduce noise in the training, as images of the same place with small visual overlap have the same impact on training as images that share more visual cues. To optimize the training process, and avoid that it stalls in local minima, most methods include hard-pair mining procedures during training~\cite{Arandjelovic2017,liu2019stochastic}. These strategies are time- and memory-wise expensive, but were meant necessary to counteract the negative effect of noisy labels on the training process. The representation power of the learned image descriptors is such that, in many cases and on large datasets (e.g. the Mapillary Street level Sequences~\cite{msls}), retrieved image lists are post-processed by heavy  re-ranking strategies~\cite{wang2022transvpr, sarlin20superglue, hausler2021patch,delg} to achieve good results. 

A step towards reducing the impact of label noise in training VPR models  was done in~\cite{GCL}, where graded image similarity labels were computed using a proxy based on approximating the field-of-view overlap of cameras. This was inspired by the frustum overlap included in a camera re-localization pipeline in~\cite{balntas2018relocnet}. We display an example of graded image similarity in Figure~\ref{fig:continuous_sim}:  a reference image with four image matches taken at different distances. The annotated image similarity $\psi$ decreases gradually as the distance increases, but there is no clear threshold between \emph{similar} and \emph{dissimilar}. In~\cite{GCL}, a novel Generalized Contrastive Loss (GLC) was also proposed to embed the graded similarity into the training process. 
However, the graded ground truth only weighs the descriptor distance in the latent space during training. This influences the gradient updates so that the descriptors of more similar images are pushed more closely to each other in the latent space. It does not ensure that the descriptor distance is an actual metric of how similar two images are.  

In this paper, we take a different direction w.r.t. existing approaches, shifting from the contrasting learning approach and recasting training VPR descriptors as a regression task, such that their distance in the latent space is a direct measure of the field-of-view overlap of the cameras in the real world, thus of the image similarity. We revamp a largely disregarded and straightforward regression-based approach, only used in a re-localization pipeline~\cite{balntas2018relocnet}, to learn VPR descriptors, and make the following contributions:
\begin{itemize}
    \item we demonstrate that regression is a powerful solution to train descriptors for VPR (especially for transformers), achieving performance comparable or superior to SoTA methods, without need of pair-mining or re-ranking strategies, while saving time and memory;
    \item we achieve high data-efficiency, i.e. models trained via regression require only a few thousands training iterations on a small set of image pairs to converge and achieve high retrieval results, and good generalization performance, differently from contrastive approaches that require multiple epochs on the training data.

\end{itemize}

The direct link between descriptor distance and image similarity imposed by our formulation results in retrieval (ranking) performance higher than or comparable to that of more sophisticated methods that use hard-pair mining for training or  re-ranking strategies. The distance in the latent space of descriptors learned by regression is indeed better representative of image similarity than that of descriptors learned via contrastive methods;




\section{Previous work}

\noindent \textbf{Visual place recognition. } Methods for visual place recognition are usually designed to solve an image retrieval task: given a query image, similar images are to be retrieved from a database (map). Traditional VPR methods build on extracting local features and combining them into holistic image descriptors, such as Bag-of-Words~\cite{bow2012,Torii-PAMI2015}, Fisher vectors~\cite{perronnin2010large} and VLAD~\cite{jegou2011VLAD}. Also, information about image sequences was used to improve performance~\cite{Milford2012}.

Current state-of-the-art approaches rely on descriptors learned via deep learning and convolutional networks (CNNs)~\cite{zhang2020visual,masone2021survey}. Initially, pre-trained networks were shown to extract effective descriptors~\cite{chen2014convolutional,lopez2019place,leyvavallina2019access}. However, training them using image pairs or triplets (depicting the same place or not) in an end-to-end fashion was demonstrated to boost the performance results largely~\cite{radenovic2018fine,leyvavallina2019caip}. The combination of a CNN backbone with a trainable VLAD layer, in NetVLAD, contributed to achieving state-of-the-art performance on several benchmarks~\cite{Arandjelovic2017}. Recently, Transformers have been deployed as backbones for VPR, to learn more powerful and effective image descriptors. TransVPR~\cite{wang2022transvpr} and CosPlace~\cite{BertonCosplace} trained a transformer backbone respectively using a triplet learning architecture and a classification-based approach. R2Former~\cite{zhu2023r2former} combined learning transformer-based descriptors and re-ranking end-to-end. These methods achieved high performance and generalization, attributable to the attention mechanism intrinsic to the transformer architecture and to the large datasets used for training (e.g. MSLS and SF-XL).



\noindent \textbf{Metric Learning for VPR. } Image descriptors for VPR are learned so that their distance is small in a latent space for images of the same place, and large for images of different places~\cite{radenovic2018fine,Gordo2017}. This approach is referred to as metric learning. Image pairs or triplets, labelled as depicting the same place or not, are used to train neural networks to compute descriptors in a metric space. 
Most popular and top-performing approaches are based on triplet network learning, e.g. NetVLAD~\cite{Arandjelovic2017} (with an architecture composed of a VGG backbone and a trainable VLAD layer). These methods use image triplets with a reference image (anchor), a positive image (depicting the same place of the anchor), and a negative image (depicting a different place). A triplet loss is then optimized to reduce the distance between the anchor and positive image descriptor, and maximize that between the anchor and negative image descriptor. Several improvements of NetVLAD were proposed, such as the fusion of patch- and global-level VLAD descriptors in PatchNetVLAD~\cite{hausler2021patch}, or training with a stochastic attraction-repulsion triplet loss function in NetVLAD-SARE~\cite{liu2019stochastic}. PointNetVLAD~\cite{angelina2018pointnetvlad} combined 3D information from PointNet~\cite{qi2017pointnet} into NetVLAD. Loss functions with weighting schemes for image pairs based on GPS distance were also explored to train NetVLAD~\cite{thoma2020soft}. 

These methods are trained using hard-pair mining to compose training batches, which requires costly computations so that the noise in data and labels is counteracted. However, the learned descriptors show some limited performance when used on larger  diverse datasets. Thus, two-stage approaches are deployed, with the top-k retrieved candidates being re-ranked using geometric verification, based on keypoint matching, either from convolutional features~\cite{hausler2021patch} or attention maps~\cite{wang2022transvpr}. Our work differs from existing metric learning approaches for VPR as we do not deploy contrastive learning, and demonstrate that a regression-based learning approach allows training (larger) models efficiently, resulting in descriptors with higher representation capabilities, and with no need for hard mining or re-ranking.

\noindent \textbf{Limitations of contrastive learning for VPR. } The common assumption to train VPR methods is that two images either depict the same place or not, in a binary way. Ground truth labels in benchmark datasets (e.g. MSLS~\cite{msls} or Pittsburgh30k~\cite{Torii-CVPR2013}) are available for image pairs,  indicating whether they show the same place or not. 
This approach does not take into account that images of the same place may share more or less visual cues, depending for example on perspective changes in the camera position. Hence, image pairs depicting the same place with large or small visual cue share are weighted in the same manner, introducing noise during training~\cite{GCL}. Computation-heavy hard-pair mining is used to select hard image pairs and triplets to avoid training stalling~\cite{Arandjelovic2017}. These methods are data-hungry, due to the necessity of overcoming the label noise problems. 

In~\cite{GCL}, a first approach to use graded  instead of binary similarity labels for image pairs was explored in combination with a Generalized Contrastive Loss (GCL) function. A relevant outcome was that hard-pair mining deemed not necessary to ensure training convergence. Larger backbones were trained efficiently, while achieving higher results than existing methods. However, the contrastive learning paradigm still introduces an artificial binarization of the problem by dividing the loss function into a positive and a negative term. 

In this work, we further exploit the concept of graded image similarity~\cite{GCL} and train VPR models by a regression task. In this way, the descriptor distance in the latent space has a direct relation with the similarity degree of images. 
The outcome is a more robust distance-based ranking of retrieved images based on similarity to the query image.

\section{Methodology}

\subsection{Architecture, optimization and training batches}
We use a siamese architecture to train the encoder of a (hybrid) visual transformer~\cite{dosovitskiy2021an} or a convolutional backbone with a GeM pool layer~\cite{radenovic2018fine}.
We optimize a Mean Squared Error loss function, which is formulated as:
\begin{equation}
\nonumber
   \mathcal{L}(x_i,x_j,\psi_{i,j})=\left \| d(\theta(x_i),\theta(x_j)) - (1-\psi_{i,j}) \right \|_2^2, 
\end{equation}

\noindent where $x_i$ and $x_j$ are the input images, which have ground truth similarity $\psi_{i,j}$, with $\theta(x_i)$ and $\theta(x_j)$ their respective descriptors computed with a model $\theta(\cdot)$. We denote $d(\cdot, \cdot)$ as a distance function in the descriptor space. We consider the Euclidean distance,  which led to better results than the Cosine distance. We L2-normalize the descriptors. 

We  do not perform hard mining to select samples to compose the training batches. We rely on the graded similarity ground truth available with the images to compose the training batches so that 50\% of the pairs have $\psi \in (0.5,1]$, 25\% with $\psi \in (0,0.5]$ and the remaining 25\% has $\psi=0$~\cite{GCL}. This ensures that a batch has pairs with approximately uniformly distributed ground truth similarity in the interval $[0,1]$. 

\subsection{Image search and retrieval}
In visual place recognition, a set of query images $Y$ taken from unknown positions are localized in an environment by comparing them with similar map images retrieved from a set $X$, for which the camera pose is known. 
We compute descriptors of the map images by $\theta(x),~\forall x \in X$, and of the query images $\theta(y),~\forall y \in Y$ using the encoder models that we train. For a given query image descriptor $\theta(y)$, image retrieval is performed by an exhaustive nearest neighbour search among the descriptors of the map set $\theta(x)\,~ \forall x \in X$. We retrieve a set of map images, which are ranked according to the closest distance of their descriptor in the latent space with respect to the query image descriptor. 
We point out that we do not address the pose estimation and camera localization tasks. Our focus is on training encoder models for visual place recognition to compute effective image descriptors and enable image retrieval to provide high-quality ranking results. The improved ranked retrieved images have the potential of boosting the performance of camera localization pipelines.

\subsection{Training data}
We train on the Mapillary Street Level Sequences (MSLS) dataset, a large-scale place recognition dataset that contains images taken in 30 cities across six continents~\cite{msls}. It includes challenging variations of camera viewpoint, season, time and illumination. The training set contains over 500k query images and 900k map images from 22 cities. The validation set consists of 19k map images and 11k query images from two cities, and the test set has 39k map images and 27k query images from six different cities.  For training, we use the FoV overlap introduced in~\cite{GCL} as ground truth image similarity.

\subsection{Evaluation data} We evaluate the trained models on the following datasets. 
\noindent\textbf{MSLS.} We use the MSLS validation set and the MSLS test set. For the latter, since the ground truth is not publicly available, we submit the predictions of our methods to the official MSLS evaluation server. Following the protocol in~\cite{msls}, two images are referenced as similar for evaluation if they are taken by cameras located within $25m$, and with less than $40^{\circ}$ of viewpoint variation.

\noindent\textbf{Pittsburgh. } It contains images recorded via Google Street View in the city of Pittsburgh, Pennsylvania, over a span of several years~\cite{Torii-PAMI2015}. We use the test set of Pittsburgh30k, which is a widely used benchmark~\cite{Arandjelovic2017}. It contains 10k map images and 7k query images, and we use it to evaluate the generalization and out-of-distribution performance of our models trained on the MSLS dataset.

\noindent\textbf{Tokyo 24/7. }
The dataset consists of 315 query images and  76k map images taken in the city of Tokyo, Japan. The dataset images show large variations of illumination, as they are taken during day and night~\cite{Torii-CVPR2013}. It is a commonly used dataset for benchmarking, explicitly designed to test robustness to changes of illumination.

\begin{table*}[!t]
\centering
\caption{Comparison to state-of-the-art methods trained on the MSLS dataset. The mark $^\star$ denotes methods that perform re-ranking. TL indicates the use of a triplet loss. We underline the best results by our methods and show the best results overall in bold.}
\label{tab:sota}
\setlength{\tabcolsep}{5pt}
\resizebox{\textwidth}{!}{%
\small
\begin{tabular}{@{}lcc@{\hskip 15pt}ccc@{\hskip 20pt}ccc@{\hskip 20pt}ccc@{\hskip 20pt}ccc@{}}
\toprule
 &  &  & \multicolumn{3}{c}{\textbf{MSLS-Val}} & \multicolumn{3}{c}{\textbf{MSLS-Test}} & \multicolumn{3}{c}{\textbf{Pitts30k}} & \multicolumn{3}{c}{\textbf{Tokyo24/7}} \\
\textbf{Encoder} & \textbf{PCA$_w$} & \textbf{Dim} & \textbf{R@1} & \textbf{R@5} & \textbf{R@10} & \textbf{R@1} & \textbf{R@5} & \textbf{R@10} & \textbf{R@1} & \textbf{R@5} & \textbf{R@10} & \textbf{R@1} & \textbf{R@5 } & \textbf{R@10} \\ \midrule
NetVLAD (TL) & N & 32768 & 44.6 & 61.1 & 66.4 & 28.8 & 44.0 & 50.7 & 40.4 & 64.5 & 74.2 & 11.4 & 24.1 & 31.4 \\
NetVLAD (TL) & Y & 4096 & 70.1 & 80.8 & 84.9 & 45.1 & 58.8 & 63.7 & 68.6 & 84.7 & 88.9 & 34.0 & 47.6 & 57.1 \\
TransVPR (TL)~\cite{wang2022transvpr} & - & - & 70.8 & 85.1 & 86.9 & 48 & 67.1 & 73.6 & 73.8 & 88.1 & 91.9 & - & - & - \\
ResNeXt-GeM-GCL~\cite{GCL} & N & 2048 & 75.5 & 86.1 & 88.5 & 56.0 & 70.8 & 75.1 & 64.0 & 81.2 & 86.6 & 37.8 & 53.6 & 62.9 \\
ResNeXt-GeM-GCL~\cite{GCL} & Y & 1024 & 80.9 & 90.7 & 92.6 & 62.3 & 76.2 & 81.1 & 79.2 & 90.4 & 93.2 & 58.1 & 74.3 & 78.1 \\
ViT-GCL & N & 768 & 71.2 & 84.9 & 88.4 & 50.0 & 67.9 & 73.9 & 70.8 & 88.0 & 91.9 & 40.3 & 60.0 & 68.9 \\
ViT-GCL & Y & 768 & 80.0 & 91.1 & 92.4 & 57.9 & 74.0 & 78.8 & 83.6 & 93.8 & 95.8 & 72.1 & 83.2 & 87.0 \\
ViT-R50-GCL & N & 768 & 69.7 & 81.6 & 85.3 & 46.8 & 62.7 & 69.5 & 70.9 & 86.7 & 91.2 & 42.9 & 62.9 & 70.5 \\
ViT-R50-GCL & Y & 768 & 78.6 & 88.1 & 90.4 & 55.9 & 71.1 & 77.2 & 82.6 & 92.8 & 95.4 & 76.2 & 86.0 & 87.3 \\
\midrule
SP-SuperGlue$^\star$~\cite{sarlin20superglue} & - & - & 78.1 & 81.9 & 84.3 & 50.6 & 56.9 & 58.3 & 87.2 & 94.8 & 96.4 & 88.2 & 90.2 & 90.2 \\
DELG$^\star$~\cite{delg} & - & - & 83.2 & 90.0 & 91.1 & 52.2 & 61.9 & 65.4 & \textbf{89.9} & \textbf{95.4} & \textbf{96.7} & \textbf{95.9} & \textbf{96.8} & \textbf{97.1} \\
Patch NetVLAD$^\star$~\cite{hausler2021patch} & Y & 4096 & 79.5 & 86.2 & 87.7 & 48.1 & 57.6 & 60.5 & 88.7 & 94.5 & 95.9 & 86.0 & 88.6 & 90.5 \\
TransVPR$^\star$~\cite{wang2022transvpr} & - & - & \textbf{86.8} & \textbf{91.2} & 92.4 & 63.9 & 74 & 77.5 & 89 & 94.9 & 96.2 & - & - & - \\\midrule

NetVLAD-MSE (ours) & N & 32768 & 72.3 & 82.7 & 85.5 & 51.5 & 64.7 & 70.7 & 74.7 & 87.2 & 90.9 & 20.7  & 41.0 & 50.8   \\
NetVLAD-MSE (ours) & Y & 4096 & 71.4 & 82.7 & 85.8 & 51.3 & 66.0 & 71.3 & 53.5 & 75.2 & 82.9 & 44.8 & 60.3 & 66.7  \\
ViT-MSE (ours) & N & 768 & 80.4 & 90.5 & 93.2 & 58.7 & 73.7 & 79.6 & 82.1 & 92.7 & 95.1 & 55.9 & 73.0 & 76.5 \\
ViT-MSE (ours) & Y & 768 & 82.4 & 90.5 & 92.3 & 60.5 & 73.8 & 78.4 & 85.1 & 94.2 & 95.9 & 71.4 & 85.4 & 89.8 \\
ViT-R50-MSE (ours) & N & 768 & 82.6 & \underline{91.1} & \textbf{\underline{93.6}} & 61.8 & 77.2 & 80.7 & 83.9 & 93.0 & 95.1 & 59.4 & 75.9 & 82.5 \\
ViT-R50-MSE (ours) & Y & 768 & \underline{84.3} & \underline{91.1} & \textbf{\underline{93.6}} & \textbf{\underline{64.4}} & \textbf{\underline{77.3}} & \textbf{\underline{81.2}} & \underline{86.4} & \underline{94.7} & \underline{96.0} & \underline{78.1} & \underline{88.9} & \underline{91.8} \\ \bottomrule
\end{tabular}%
}
\end{table*}

\subsection{Implementation details}
We trained our models for one epoch (520k image pairs) using a Stochastic Gradient Descent (SGD) optimizer with an initial learning rate equal to $0.1$. For the training of models with the Generalized Contrastive Loss, we decrease the learning rate by a factor of $10^{-1}$ every 250k iterations. For the optimization with the Mean Squared Error loss, we keep the learning rate constant at $0.1$ for the whole training. All our experiments were carried out using PyTorch.
The models and training code will be publicly available.
\section{Experiments and results}

\subsection{Results}
\noindent\textbf{Comparison with state-of-the-art. }
We compared the results of our models with state-of-the-art approaches, considering both methods that perform only retrieval (i.e. NetVLAD~\cite{Arandjelovic2017}, TransVPR w/o re-ranking~\cite{wang2022transvpr}, GCL~\cite{GCL}) and  methods that apply re-ranking strategies to improve the list of retrieved images (i.e. PatchNetVLAD~\cite{hausler2021patch}, SP-SuperGlue~\cite{sarlin20superglue}, DELG~\cite{delg} and TransVPR with attention-based re-ranking~\cite{wang2022transvpr}). We compute the performance in terms of recall rate at k (R@k), typically used in VPR.  All considered models are trained on the MSLS dataset~\cite{msls}. We test their generalization abilities on the Pittsburgh30k~\cite{Arandjelovic2017} and the Tokyo24/7~\cite{Arandjelovic2017} datasets. Although we do not do re-ranking, our results compare in many cases favourably or on par with re-ranking approaches. This highlights the quality and effectiveness of the descriptors that we learn by regressing image similarity in the form of field-of-view overlap. We summarize the results in Table~\ref{tab:sota}.

We demonstrate that a hybrid transformer architecture (ViT-R50~\cite{dosovitskiy2021an}), trained for a single epoch by optimizing a straightforward regression loss function outperforms methods with more complex algorithms (e.g. re-ranking of the retrieved results) and training strategies (e.g. hard-pair mining for training), such as NetVLAD~\cite{Arandjelovic2017} and TransVPR~\cite{wang2022transvpr}. We indeed achieved higher results on the MSLS, Pittsburgh30k and Tokyo24/7 datasets than all methods that perform retrieval only based on descriptor distance ranking. We achieve in some cases higher results than methods that also apply re-ranking (R@5 3.3\% higher than TransVPR on the MSLS test set), or otherwise comparable and very competitive results (e.g. R@5 0.1\% lower than TransVPR on MSLS val, or 0.7\% lower than DELG on Pittsburgh30k). 

The other method that relies on graded similarity ground truth, namely GCL, achieves lower results than our models trained using regression. This highlights the fact that our regression-based approach further exploits the information provided by the graded ground truth, and result in descriptors that are more suited for image retrieval and place recognition.

 \begin{figure*}[!t]
    \centering
    \scriptsize
    \hspace{-5mm}
    \subfloat[]{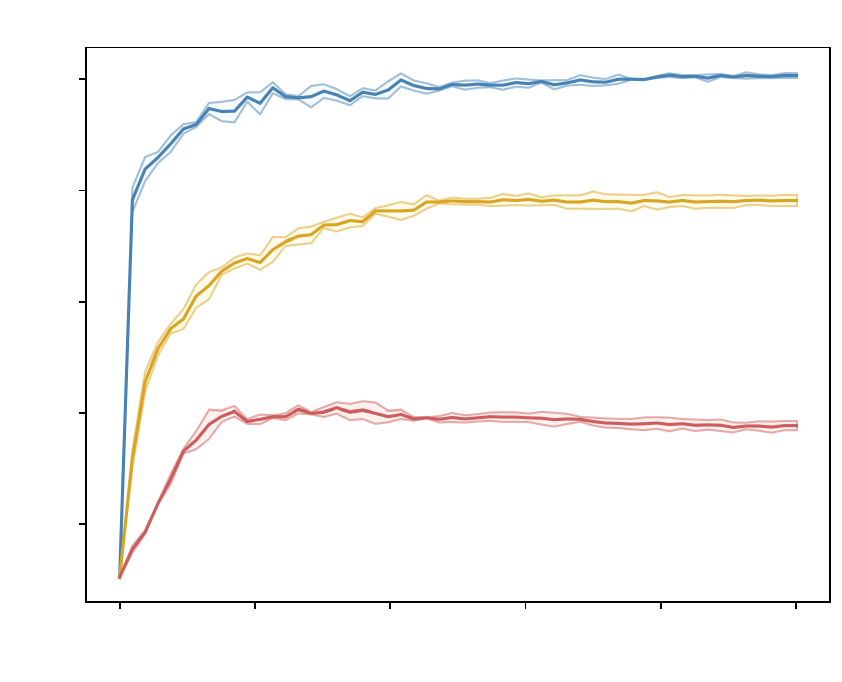}
    ~
    \subfloat[]{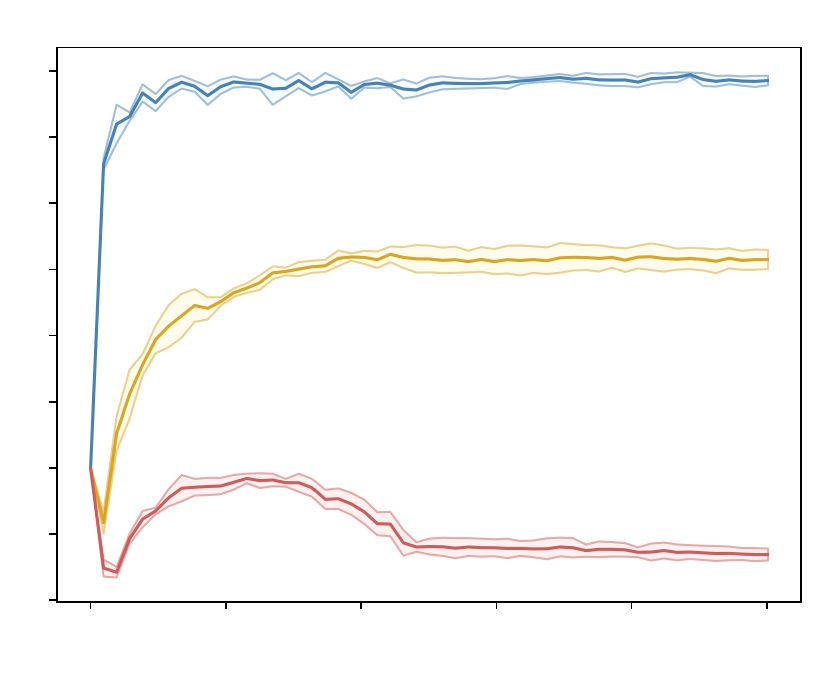}
    ~
    \subfloat[]{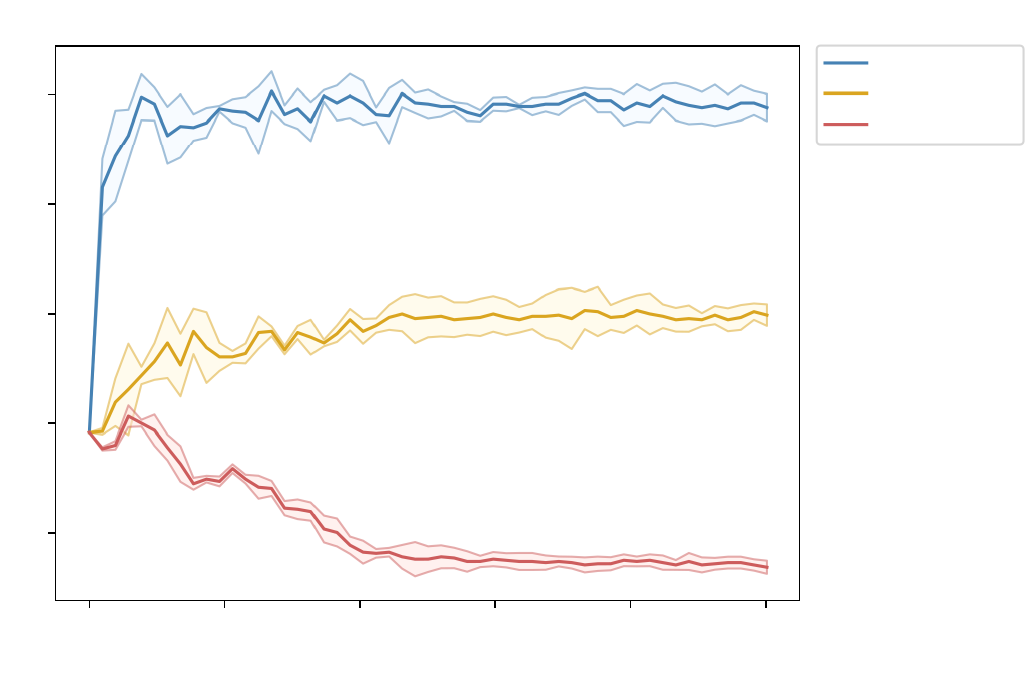} 
    \caption{Retrieval performance every 10k training iterations on the (a) MSLS-val, (b) Pitts30k, and (c) Tokyo24/7 for the same ViT-R50 encoder trained with CL, GCL, and MSE loss functions. We ran each experiment three times and report the average, minimum and maximum R@5.}
    \label{fig:dataefficiency}
\end{figure*}

\noindent\textbf{Data-efficiency. }
We study the data-efficiency of the proposed method, namely the ability to exploit few samples to train highly-performing methods in low-data regimes. We show that the proposed regression-based training requires only a few thousand image pairs (thus training iterations) to learn very effective and robust descriptors. 
In Fig.~\ref{fig:dataefficiency}, we report the results (R@5) achieved at intermediate steps while training on the MSLS dataset. We evaluate the data-efficiency of a hybrid transformer trained using the MSE loss and compare the results to those of the same backbone trained using the GCL and the binary Contrastive loss (CL). We test a model snapshot every 10k iterations (i.e. after seeing 10k training pairs) on the MSLS validation set. We also test the generalization of the model snapshots to Pittsburgh30k and Tokyo24/7.

The regression-based training is able to very quickly learn robust and generalizable image descriptors for image retrieval and VPR. It uses only a few thousand image pairs (no need to complete a single epoch) to achieve very high results, superior to those of the GCL and CL training. The training regime of the MSE loss stabilizes very quickly to retrieval performance that is much higher than other methods. This has implications on several aspects, e.g. saving unnecessary computations and energy needed to train models with complex hard-pair mining for a longer time, quickly obtaining very powerful descriptors that do not require expensive re-ranking algorithms and allowing for hyperparameter optimization or model search that would not be feasible with existing models.
It is worth pointing out that GCL is the only other method that uses data efficiently and can exploit graded image similarity ground truth. State-of-the-art methods require several training epochs and days to converge~\cite{GCL}.

\noindent\textbf{Quality of ranked retrieval results.}
We measured the quality of the retrieval ranking of our models using three different metrics. We compute the Recall@5 (R@5), which considers the place of a query image as correctly recognized if there is at least one true positive among the top-5 retrieved database images. This is a standard metric in visual place recognition and image retrieval but does not take into account the position of those true positives in the ranking. We compute the top-5 Mean Reciprocal Ranking (MRR@5) to measure the position in which the first positive hit appears in the top-5 retrieved images. It is equal to $1$ if the first retrieved image is a true positive, and f.i. is equal to $4/5$ if the first positive hit is in the second position, until the score equals to 0 if there is no positive retrieved among the 5 first results. When training a model with a ViT-R50 using a regression loss, not only the recall is higher, but correct images are placed earlier in the ranked list than with the counterpart model trained using a contrastive approach with graded similarity labels (see Table~\ref{tab:ranking}). This also applies in out-of-distribution tests on the Pitts30k and Tokyo24/7 datasets, demonstrating generalization of ranking capabilities. We compare with the GCL as it is the only method trained using graded similarity at large-scale.

\begin{table}[!t]
\centering
\caption{Comparison of ranked retrieval results of backbones trained with GCL~\cite{GCL} and MSE (ours) loss.}
\label{tab:ranking}
\footnotesize
\begin{tabular}{@{}lccc@{}}
\toprule
\textbf{Dataset} & \textbf{Metric}                            & \textbf{ViT-R50-GCL} & \textbf{ViT-R50-MSE}  \\ \midrule
\multirow{3}{*}{MSLS-val}  & R@5 & 81.6                 & \textbf{91.1}         \\
                           & MRR-5                   & 0.4399               & \textbf{0.5433}       \\
                           & KLDiv                   & \num{14e-4}          & \textbf{\num{1e-4}}   \\ \midrule
\multirow{3}{*}{Pitts30k}  & R@5                     & 86.7                 & \textbf{93.0}         \\
                           & MRR@5                  & 0.5024               & \textbf{0.6351}       \\
                           & KLDiv                  & \num{30.6e-3}        & \textbf{\num{5.7e-3}} \\ \midrule
\multirow{3}{*}{Tokyo24/7} & R@5                     & 62.9                 & \textbf{75.9}         \\
                           & MRR@5                   & 0.3226               & \textbf{0.4505}       \\
                           & KLDiv                  & \num{37e-4}          & \textbf{\num{8e-4}}   \\ \bottomrule
\end{tabular}%


\end{table}

\begin{table*}[!t]
\caption{Ablation study of encoders and PCA. All models are trained on the MSLS dataset with an MSE loss.}
\label{tab:ablation}
\resizebox{\textwidth}{!}{%
\footnotesize
\setlength{\tabcolsep}{5pt}
\begin{tabular}{@{}lcc@{\hskip 20pt}ccc@{\hskip 20pt}ccc@{\hskip 20pt}ccc@{\hskip 20pt}ccc@{}}
\toprule
\multirow{2}{*}{\textbf{Encoder}} & \multirow{2}{*}{\textbf{PCA$_w$}} & \multirow{2}{*}{\textbf{Dim}} & \multicolumn{3}{c}{\textbf{MSLS-val}} & \multicolumn{3}{c}{\textbf{MSLS-test}} & \multicolumn{3}{c}{\textbf{Pitts30k}} & \multicolumn{3}{c}{\textbf{Tokyo24/7}} \\
 &  &  & \textbf{R@1} & \textbf{R@5} & \textbf{R@10} & \textbf{R@1} & \textbf{R@5} & \textbf{R@10} & \textbf{R@1} & \textbf{R@5} & \textbf{R@10} & \textbf{R@1} & \textbf{R@5} & \textbf{R@10} \\ \midrule
NetVLAD & N & 32768 & 72.3 & 82.7 & 85.5 & 51.5 & 64.7 & 70.7 & 74.7 & 87.2 & 90.9 & 20.7  & 41.0 & 50.8 \\
 & Y & 4096 & 71.4 & 82.7 & 85.8 & 51.3 & 66.0 & 71.3 & 53.5 & 75.2 & 82.9 & 44.8 & 60.3 & 66.7 \\
VGG16-GeM & N & 512 & 69.3 & 81.1 & 84.2 & 43.9 & 57.3 & 62.6 & 66.1 & 84.0 & 89.2 & 33.3 & 57.1 & 64.8 \\
 & Y & 256 & 65.4 & 78 & 81.6 & 38.6 & 54.2 & 60.8 & 62.3 & 80.6 & 86.4 & 28.3 & 47.0 & 52.4 \\
ResNeXT-GeM & N & 2048 & 78.8 & 88.6 & 90.7 & 58.5 & 73.1 & 78.6 & 71.9 & 86.8 & 91.3 & 41.6 & 63.5 & 71.4 \\
 & Y & 1024 & 81.1 & 90.8 & 92.4 & 59.7 & 73.1 & 77.4 & 74.3 & 87.6 & 91.5 & 51.8 & 69.7 & 78.4 \\
ViT & N & 768 & 80.4 & 90.5 & 93.2 & 58.7 & 73.7 & 79.6 & 82.1 & 92.7 & 95.1 & 55.9 & 73.0 & 76.5 \\
 & Y & 768 & 82.4 & 90.5 & 92.3 & 60.5 & 73.8 & 78.4 & 85.1 & 94.2 & 95.9 & 71.4 & 85.4 & 89.8 \\
ViT-R50 & N & 768 & 82.6 & \textbf{91.1} & \textbf{93.6} & 61.8 & 77.2 & 80.7 & 83.9 & 93.0 & 95.1 & 59.4 & 75.9 & 82.5 \\
 & Y & 768 & \textbf{84.3} & \textbf{91.1} & \textbf{93.6} & \textbf{64.4} & \textbf{77.3} & \textbf{81.2} & \textbf{86.4} & \textbf{94.7} & \textbf{96.0} & \textbf{78.1} & \textbf{88.9} & \textbf{91.8} \\
 \bottomrule
\end{tabular}%
}

\end{table*}

Furthermore, we compute the Kullback-Leibler divergence between the distribution of query-map pairwise Euclidean distance and the annotated similarity ground truth of image pairs. For MSLS, the ground truth is the FoV overlap~\cite{GCL}. For Pitts30k and Tokyo24/7 datasets, we consider the binary ground truth publicly available. The Kullback-Leibler divergence measures how different two distributions are. As such, a lower value indicates a higher similarity between distributions. The KL of the MSE-trained model is at least one order of magnitude smaller than that of its GCL counterpart (Table~\ref{tab:ranking}). The MSE-trained descriptors are  thus better suited for VPR, as their distance in the  latent space is a more robust measure of image visual similarity and ranking. 

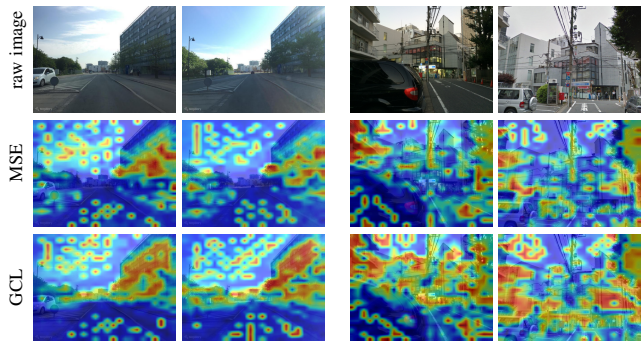
\begin{figure}[!t]
    \centering
\begingroup%
  \makeatletter%
  \providecommand\color[2][]{%
    \errmessage{(Inkscape) Color is used for the text in Inkscape, but the package 'color.sty' is not loaded}%
    \renewcommand\color[2][]{}%
  }%
  \providecommand\transparent[1]{%
    \errmessage{(Inkscape) Transparency is used (non-zero) for the text in Inkscape, but the package 'transparent.sty' is not loaded}%
    \renewcommand\transparent[1]{}%
  }%
  \providecommand\rotatebox[2]{#2}%
  \newcommand*\fsize{\dimexpr\f@size pt\relax}%
  \newcommand*\lineheight[1]{\fontsize{\fsize}{#1\fsize}\selectfont}%
  \ifx\svgwidth\undefined%
    \setlength{\unitlength}{240bp}%
    \ifx\svgscale\undefined%
      \relax%
    \else%
      \setlength{\unitlength}{\unitlength * \real{\svgscale}}%
    \fi%
  \else%
    \setlength{\unitlength}{\svgwidth}%
  \fi%
  \global\let\svgwidth\undefined%
  \global\let\svgscale\undefined%
  \makeatother%
  \begin{picture}(1,0.53986788)%
    \lineheight{1}%
    \scriptsize
    \setlength\tabcolsep{0pt}%
    \put(0,0){\includegraphics[width=\unitlength,page=1]{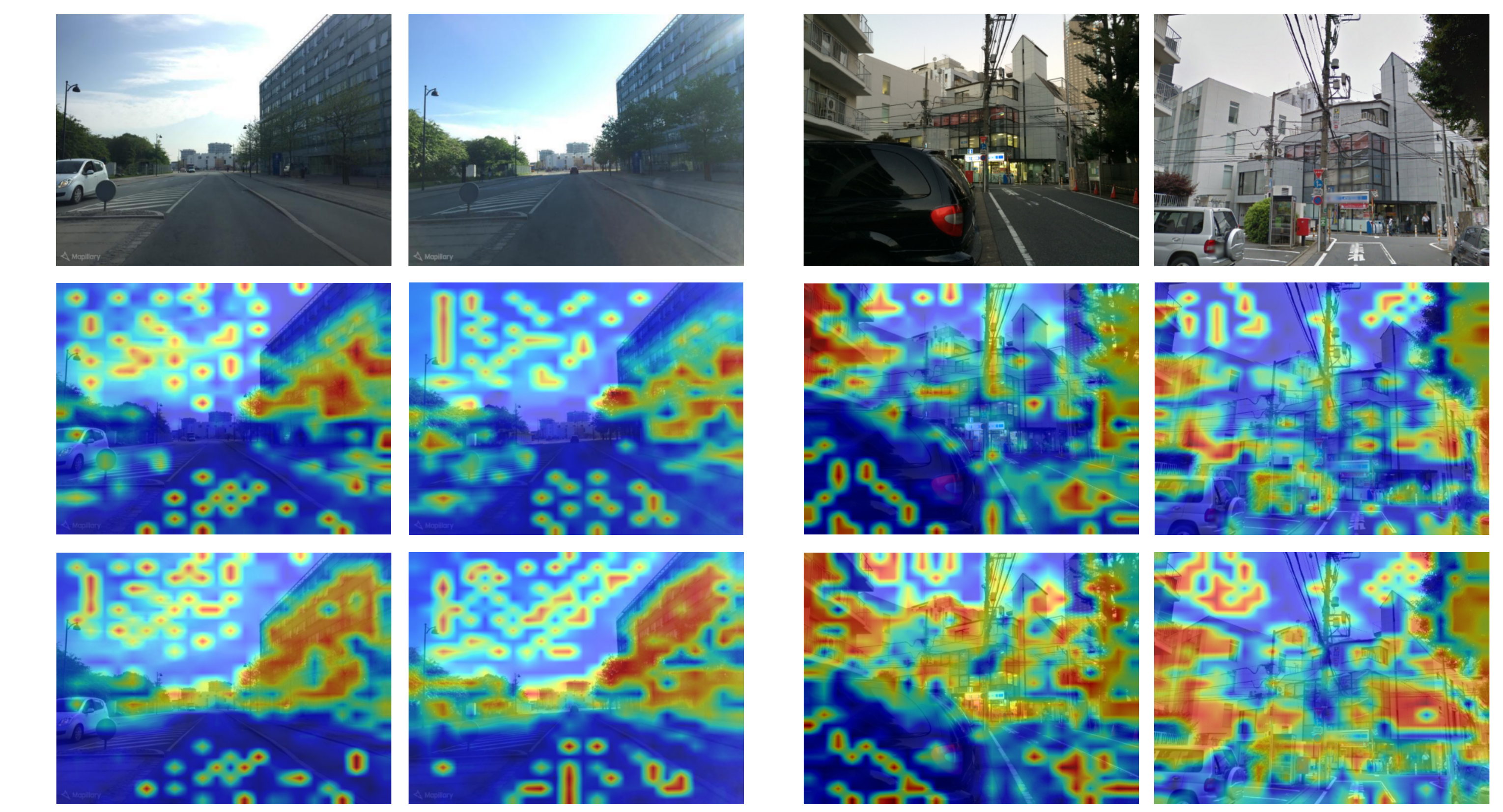}}%
    \put(0.02,0.4){\rotatebox{90}{\makebox(0,0)[lt]{\smash{\begin{tabular}[t]{l}raw image\end{tabular}}}}}%
    \put(0.02,0.255){\rotatebox{90}{\makebox(0,0)[lt]{\smash{\begin{tabular}[t]{l}MSE\end{tabular}}}}}%
    \put(0.02,0.06){\rotatebox{90}{\makebox(0,0)[lt]{\smash{\begin{tabular}[t]{l}GCL\end{tabular}}}}}%
  \end{picture}%
\endgroup%

    \caption{Example attention maps on the last layer of ViT-R50-MSe and ViT-R50-GCL models for pairs of similar images from the MSLS validation dataset (columns 1-2), and the Tokyo24/7 dataset (columns 3-4).}
    \label{fig:attention}
\end{figure}

\noindent\textbf{Attention maps. }
We show example attention maps of our regression-trained transformer in comparison with those of its counterpart model trained with a GCL function in Fig.~\ref{fig:attention}. Although both models tend to have attention maps with focus on relevant parts of the images, the model trained with MSE reacts less to non-permanent features, such as the car present in the left images (both MSLS and Tokyo24/7). This supports the evidence that the regression-trained model performs better than contrastive-based approaches, as it focuses on structural shared visual clues in images.

\begin{figure*}[!t]
\subfloat{
\centering
\hspace{2.0cm}
\begin{minipage}[c]{0.9\textwidth}
\begin{tikzpicture}[thick, scale=.8, every node/.style={scale=1}] 
\begin{axis}[%
hide axis,
xmin=20,
xmax=50,
ymin=0,
ymax=0.4,
legend style={draw=white!15!black,legend cell align=left,legend columns=4}, height=4.75cm, width=1\textwidth
]
\addlegendimage{green,thick, dashed, mark=square, mark size=2pt, only marks}
\addlegendentry{VGG16-GeM-GCL};
\addlegendimage{magenta,thick, dashed, mark=o, mark size=2pt, only marks}
\addlegendentry{ResNeXt-GeM-GCL};
\addlegendimage{blue,thick, dashed, mark=star, mark size=2pt, only marks}
\addlegendentry{ViT-GCL};
\addlegendimage{cyan,thick, dashed, mark=triangle, mark size=2pt, only marks}
\addlegendentry{ViT-R50-GCL};

\addlegendimage{green,thick, dashed, mark=square, mark size=2pt}
\addlegendentry{VGG16-GeM-GCL-PCA$_w$};
\addlegendimage{magenta,thick, dashed, mark=o, mark size=2pt}
\addlegendentry{ResNeXt-GeM-GCL-PCA$_w$};
\addlegendimage{blue,thick, dashed, mark=star, mark size=2pt}
\addlegendentry{ViT-GCL-PCA$_w$};
\addlegendimage{cyan,thick, dashed, mark=triangle, mark size=2pt}
\addlegendentry{ViT-R50-GCL-PCA$_w$};

\addlegendimage{green,thick, solid, mark=pentagon, mark size=2pt, only marks}
\addlegendentry{VGG16-GeM-MSE};
\addlegendimage{magenta,thick, solid, mark=oplus, mark size=2pt, only marks}
\addlegendentry{ResNeXt-GeM-MSE};
\addlegendimage{blue,thick, solid, mark=10-pointed star, mark size=2pt, only marks}
\addlegendentry{ViT-MSE};
\addlegendimage{cyan,thick, solid, mark=diamond, mark size=2pt, only marks}
\addlegendentry{ViT-R50-MSE};

\addlegendimage{green,thick, solid, mark=pentagon, mark size=2pt}
\addlegendentry{VGG16-GeM-MSE-PCA$_w$};
\addlegendimage{magenta,thick, solid, mark=oplus, mark size=2pt}
\addlegendentry{ResNeXt-GeM-MSE-PCA$_w$};
\addlegendimage{blue,thick, solid, mark=10-pointed star, mark size=2pt}
\addlegendentry{ViT-MSE-PCA$_w$};
\addlegendimage{cyan,thick, solid, mark=diamond, mark size=2pt}
\addlegendentry{ViT-R50-MSE-PCA$_w$};
    \end{axis}
    \end{tikzpicture} 
    \end{minipage}
}

\vspace{-35pt}
\setcounter{subfigure}{0}

    \subfloat[MSLS validation set]{
\begin{tikzpicture}[thick, scale=.41, every node/.style={scale=1.25}]
\begin{axis}
[xlabel=Dimensions,
ylabel=Recall@5 (\%), ylabel style={at={(axis description cs:0.05,0.5)}},grid, xmode=log,log basis x={2}, xtick={32,64,128,256, 512,1024,2048}, xticklabels={32,64,128,256, 512,1024,2048}, width=.6\textwidth
]
\addplot[green,thick, dashed, mark=square, mark size=3.5pt] table [x=d, y=r, col sep=comma] {data/whitening/MSLS_val/MSLS_vgg16_GeM_GCL_nowhiten_toplot.txt};
\addplot[green, mark=square, dashed, thick, mark size=3.5pt] table [x=d, y=r, col sep=comma] {data/whitening/MSLS_val/MSLS_vgg16_GeM_GCL_whiten_toplot.txt};
\addplot[green,thick, mark=pentagon, mark size=3.5pt] table [x=d, y=r, col sep=comma] {data/whitening/MSLS_val/MSLS_vgg16_GeM_mse_nowhiten_toplot.txt};
\addplot[green, mark=pentagon, thick, mark size=3.5pt] table [x=d, y=r, col sep=comma] {data/whitening/MSLS_val/MSLS_vgg16_GeM_mse_whiten_toplot.txt};

\addplot[magenta,thick, mark=o,dashed, mark size=3.5pt] table [x=d, y=r, col sep=comma] {data/whitening/MSLS_val/MSLS_resnext_GeM_GCL_nowhiten_toplot.txt};
\addplot[magenta, mark=o, dashed, thick, mark size=3.5pt] table [x=d, y=r, col sep=comma] {data/whitening/MSLS_val/MSLS_resnext_GeM_GCL_whiten_toplot.txt};
\addplot[magenta,thick, mark=oplus, mark size=3.5pt] table [x=d, y=r, col sep=comma] {data/whitening/MSLS_val/MSLS_resnext_GeM_mse_best_nowhiten_toplot.txt};
\addplot[magenta, mark=oplus, thick, mark size=3.5pt] table [x=d, y=r, col sep=comma] {data/whitening/MSLS_val/MSLS_resnext_GeM_mse_best_whiten_toplot.txt};

\addplot[blue,thick, mark=star,dashed, mark size=3.5pt] table [x=d, y=r, col sep=comma] {data/whitening/MSLS_val/MSLS_vitb16_GCL_best_nowhiten_toplot.txt};
\addplot[blue, mark=star, dashed, thick, mark size=3.5pt] table [x=d, y=r, col sep=comma] {data/whitening/MSLS_val/MSLS_vitb16_GCL_best_whiten_toplot.txt};
\addplot[blue,thick, mark=10-pointed star, mark size=3.5pt] table [x=d, y=r, col sep=comma] {data/whitening/MSLS_val/MSLS_vitb16_mse_best_nowhiten_toplot.txt};
\addplot[blue, mark=10-pointed star, thick, mark size=3.5pt] table [x=d, y=r, col sep=comma] {data/whitening/MSLS_val/MSLS_vitb16_mse_best_whiten_toplot.txt};

\addplot[cyan,thick, mark=triangle,dashed, mark size=3.5pt] table [x=d, y=r, col sep=comma] {data/whitening/MSLS_val/MSLS_vit_GCL_best_nowhiten_toplot.txt};
\addplot[cyan, mark=triangle, dashed, thick, mark size=3.5pt] table [x=d, y=r, col sep=comma] {data/whitening/MSLS_val/MSLS_vit_GCL_best_whiten_toplot.txt};
\addplot[cyan,thick, mark=diamond, mark size=3.5pt] table [x=d, y=r, col sep=comma] {data/whitening/MSLS_val/MSLS_vit_mse_best_nowhiten_toplot.txt};
\addplot[cyan, mark=diamond, thick, mark size=3.5pt] table [x=d, y=r, col sep=comma] {data/whitening/MSLS_val/MSLS_vit_mse_best_whiten_toplot.txt};

\end{axis}
 \label{fig:whitening_msls_val}
\end{tikzpicture}
   
}
   \subfloat[MSLS test set]{
\begin{tikzpicture}[thick, scale=.41, every node/.style={scale=1.25}]
\begin{axis}
[xlabel=Dimensions, ylabel style={at={(axis description cs:-0.08,0.5)}},grid, xmode=log,log basis x={2}, xtick={32,64,128,256, 512,1024,2048}, xticklabels={32,64,128,256, 512,1024,2048}, width=.6\textwidth
]
\addplot[green,thick, dashed, mark=square, mark size=3.5pt] table [x=d, y=r, col sep=comma] {data/whitening/MSLS_test/MSLS_vgg16_GeM_GCL_nowhiten_toplot.txt};
\addplot[green, mark=square, dashed, thick, mark size=3.5pt] table [x=d, y=r, col sep=comma] {data/whitening/MSLS_test/MSLS_vgg16_GeM_GCL_whiten_toplot.txt};
\addplot[green,thick, mark=pentagon, mark size=3.5pt] table [x=d, y=r, col sep=comma] {data/whitening/MSLS_test/MSLS_vgg16_GeM_mse_nowhiten_toplot.txt};
\addplot[green, mark=pentagon, thick, mark size=3.5pt] table [x=d, y=r, col sep=comma] {data/whitening/MSLS_test/MSLS_vgg16_GeM_mse_whiten_toplot.txt};

\addplot[magenta,thick,dashed, mark=o, mark size=3.5pt] table [x=d, y=r, col sep=comma] {data/whitening/MSLS_test/MSLS_resnext_GeM_GCL_nowhiten_toplot.txt};
\addplot[magenta, mark=o, dashed, thick, mark size=3.5pt] table [x=d, y=r, col sep=comma] {data/whitening/MSLS_test/MSLS_resnext_GeM_GCL_whiten_toplot.txt};
\addplot[magenta,thick, mark=oplus, mark size=3.5pt] table [x=d, y=r, col sep=comma] {data/whitening/MSLS_test/MSLS_resnext_GeM_mse_best_nowhiten_toplot.txt};
\addplot[magenta, mark=oplus, thick, mark size=3.5pt] table [x=d, y=r, col sep=comma] {data/whitening/MSLS_test/MSLS_resnext_GeM_mse_best_whiten_toplot.txt};

\addplot[blue,thick, mark=star,dashed, mark size=3.5pt] table [x=d, y=r, col sep=comma] {data/whitening/MSLS_test/MSLS_vitb16_GCL_best_nowhiten_toplot.txt};
\addplot[blue, mark=star, dashed, thick, mark size=3.5pt] table [x=d, y=r, col sep=comma] {data/whitening/MSLS_test/MSLS_vitb16_GCL_best_whiten_toplot.txt};
\addplot[blue,thick, mark=10-pointed star, mark size=3.5pt] table [x=d, y=r, col sep=comma] {data/whitening/MSLS_test/MSLS_vitb16_mse_best_nowhiten_toplot.txt};
\addplot[blue, mark=10-pointed star, thick, mark size=3.5pt] table [x=d, y=r, col sep=comma] {data/whitening/MSLS_test/MSLS_vitb16_mse_best_whiten_toplot.txt};

\addplot[cyan,thick, mark=triangle,dashed, mark size=3.5pt] table [x=d, y=r, col sep=comma] {data/whitening/MSLS_test/MSLS_vit_GCL_best_nowhiten_toplot.txt};
\addplot[cyan, mark=triangle, dashed, thick, mark size=3.5pt] table [x=d, y=r, col sep=comma] {data/whitening/MSLS_test/MSLS_vit_GCL_best_whiten_toplot.txt};
\addplot[cyan,thick, mark=diamond, mark size=3.5pt] table [x=d, y=r, col sep=comma] {data/whitening/MSLS_test/MSLS_vit_mse_best_nowhiten_toplot.txt};
\addplot[cyan, mark=diamond, thick, mark size=3.5pt] table [x=d, y=r, col sep=comma] {data/whitening/MSLS_test/MSLS_vit_mse_best_whiten_toplot.txt};

\end{axis}
 \label{fig:whitening_msls_test}
\end{tikzpicture}  
}
\subfloat[pitts30k test set]{
\begin{tikzpicture}[thick, scale=.41, every node/.style={scale=1.25}]
\begin{axis}
[xlabel=Dimensions,,grid, xmode=log,log basis x={2}, xtick={32,64,128,256, 512,1024,2048}, xticklabels={32,64,128,256, 512,1024,2048}, width=.6\textwidth
]
\addplot[green,thick, dashed, mark=square, mark size=3.5pt] table [x=d, y=r, col sep=comma] {data/whitening/pitts30k/MSLS_vgg16_GeM_GCL_nowhiten_toplot.txt};
\addplot[green, mark=square, dashed, thick, mark size=3.5pt] table [x=d, y=r, col sep=comma] {data/whitening/pitts30k/MSLS_vgg16_GeM_GCL_whiten_toplot.txt};
\addplot[green,thick, mark=pentagon, mark size=3.5pt] table [x=d, y=r, col sep=comma] {data/whitening/pitts30k/MSLS_vgg16_GeM_mse_nowhiten_toplot.txt};
\addplot[green, mark=pentagon, thick, mark size=3.5pt] table [x=d, y=r, col sep=comma] {data/whitening/pitts30k/MSLS_vgg16_GeM_mse_whiten_toplot.txt};

\addplot[magenta,thick, mark=o,dashed, mark size=3.5pt] table [x=d, y=r, col sep=comma] {data/whitening/pitts30k/MSLS_resnext_GeM_GCL_nowhiten_toplot.txt};
\addplot[magenta, mark=o, dashed, thick, mark size=3.5pt] table [x=d, y=r, col sep=comma] {data/whitening/pitts30k/MSLS_resnext_GeM_GCL_whiten_toplot.txt};
\addplot[magenta,thick, mark=oplus, mark size=3.5pt] table [x=d, y=r, col sep=comma] {data/whitening/pitts30k/MSLS_resnext_GeM_mse_best_nowhiten_toplot.txt};
\addplot[magenta, mark=oplus, thick, mark size=3.5pt] table [x=d, y=r, col sep=comma] {data/whitening/pitts30k/MSLS_resnext_GeM_mse_best_whiten_toplot.txt};

\addplot[blue,thick, mark=star,dashed, mark size=3.5pt] table [x=d, y=r, col sep=comma] {data/whitening/pitts30k/MSLS_vitb16_GCL_best_nowhiten_toplot.txt};
\addplot[blue, mark=star, dashed, thick, mark size=3.5pt] table [x=d, y=r, col sep=comma] {data/whitening/pitts30k/MSLS_vitb16_GCL_best_whiten_toplot.txt};
\addplot[blue,thick, mark=10-pointed star, mark size=3.5pt] table [x=d, y=r, col sep=comma] {data/whitening/pitts30k/MSLS_vitb16_mse_best_nowhiten_toplot.txt};
\addplot[blue, mark=10-pointed star, thick, mark size=3.5pt] table [x=d, y=r, col sep=comma] {data/whitening/pitts30k/MSLS_vitb16_mse_best_whiten_toplot.txt};

\addplot[cyan,thick, mark=triangle,dashed, mark size=3.5pt] table [x=d, y=r, col sep=comma] {data/whitening/pitts30k/MSLS_vit_GCL_best_nowhiten_toplot.txt};
\addplot[cyan, mark=triangle, dashed, thick, mark size=3.5pt] table [x=d, y=r, col sep=comma] {data/whitening/pitts30k/MSLS_vit_GCL_best_whiten_toplot.txt};
\addplot[cyan,thick, mark=diamond, mark size=3.5pt] table [x=d, y=r, col sep=comma] {data/whitening/pitts30k/MSLS_vit_mse_best_nowhiten_toplot.txt};
\addplot[cyan, mark=diamond, thick, mark size=3.5pt] table [x=d, y=r, col sep=comma] {data/whitening/pitts30k/MSLS_vit_mse_best_whiten_toplot.txt};

\end{axis}
 \label{fig:whitening_pitts30k}
\end{tikzpicture}
}
\subfloat[Tokyo 24/7]{
\begin{tikzpicture}[thick, scale=.41, every node/.style={scale=1.25}]
\begin{axis}
[xlabel=Dimensions, ylabel style={at={(axis description cs:-0.08,0.5)}},grid, xmode=log,log basis x={2}, xtick={32,64,128,256, 512,1024,2048}, xticklabels={32,64,128,256, 512,1024,2048}, width=.6\textwidth
]
\addplot[green,thick, dashed, mark=square, mark size=3.5pt] table [x=d, y=r, col sep=comma] {data/whitening/tokyo247/MSLS_vgg16_GeM_GCL_nowhiten_toplot.txt};
\addplot[green, mark=square, dashed, thick, mark size=3.5pt] table [x=d, y=r, col sep=comma] {data/whitening/tokyo247/MSLS_vgg16_GeM_GCL_whiten_toplot.txt};
\addplot[green,thick, mark=pentagon, mark size=3.5pt] table [x=d, y=r, col sep=comma] {data/whitening/tokyo247/MSLS_vgg16_GeM_mse_nowhiten_toplot.txt};
\addplot[green, mark=pentagon, thick, mark size=3.5pt] table [x=d, y=r, col sep=comma] {data/whitening/tokyo247/MSLS_vgg16_GeM_mse_whiten_toplot.txt};

\addplot[magenta,thick,dashed, mark=o, mark size=3.5pt] table [x=d, y=r, col sep=comma] {data/whitening/tokyo247/MSLS_resnext_GeM_GCL_nowhiten_toplot.txt};
\addplot[magenta, mark=o, dashed, thick, mark size=3.5pt] table [x=d, y=r, col sep=comma] {data/whitening/tokyo247/MSLS_resnext_GeM_GCL_whiten_toplot.txt};
\addplot[magenta,thick, mark=oplus, mark size=3.5pt] table [x=d, y=r, col sep=comma] {data/whitening/tokyo247/MSLS_resnext_GeM_mse_best_nowhiten_toplot.txt};
\addplot[magenta, mark=oplus, thick, mark size=3.5pt] table [x=d, y=r, col sep=comma] {data/whitening/tokyo247/MSLS_resnext_GeM_mse_best_whiten_toplot.txt};

\addplot[blue,thick, mark=star,dashed, mark size=3.5pt] table [x=d, y=r, col sep=comma] {data/whitening/tokyo247/MSLS_vitb16_GCL_best_nowhiten_toplot.txt};
\addplot[blue, mark=star, dashed, thick, mark size=3.5pt] table [x=d, y=r, col sep=comma] {data/whitening/tokyo247/MSLS_vitb16_GCL_best_whiten_toplot.txt};
\addplot[blue,thick, mark=10-pointed star, mark size=3.5pt] table [x=d, y=r, col sep=comma] {data/whitening/tokyo247/MSLS_vitb16_mse_best_nowhiten_toplot.txt};
\addplot[blue, mark=10-pointed star, thick, mark size=3.5pt] table [x=d, y=r, col sep=comma] {data/whitening/tokyo247/MSLS_vitb16_mse_best_whiten_toplot.txt};

\addplot[cyan,thick,dashed, mark=triangle, mark size=3.5pt] table [x=d, y=r, col sep=comma] {data/whitening/tokyo247/MSLS_vit_GCL_best_nowhiten_toplot.txt};
\addplot[cyan, mark=triangle, dashed, thick, mark size=3.5pt] table [x=d, y=r, col sep=comma] {data/whitening/tokyo247/MSLS_vit_GCL_best_whiten_toplot.txt};
\addplot[cyan,thick, mark=diamond, mark size=3.5pt] table [x=d, y=r, col sep=comma] {data/whitening/tokyo247/MSLS_vit_mse_best_nowhiten_toplot.txt};
\addplot[cyan, mark=diamond, thick, mark size=3.5pt] table [x=d, y=r, col sep=comma] {data/whitening/tokyo247/MSLS_vit_mse_best_whiten_toplot.txt};

\end{axis}
\label{fig:whitening_tokyo247}
\end{tikzpicture}

}

\caption{Results obtained on the MSLS validation, MSLS test, Pittsburgh 30k and Tokyo 24/7 datasets by MSE-trained models with and without PCA whitening. Reducing the dimensionality of the descriptors and applying the whitening transform contribute to an increase of the retrieval performance.}
\label{fig:whitening}
\end{figure*}
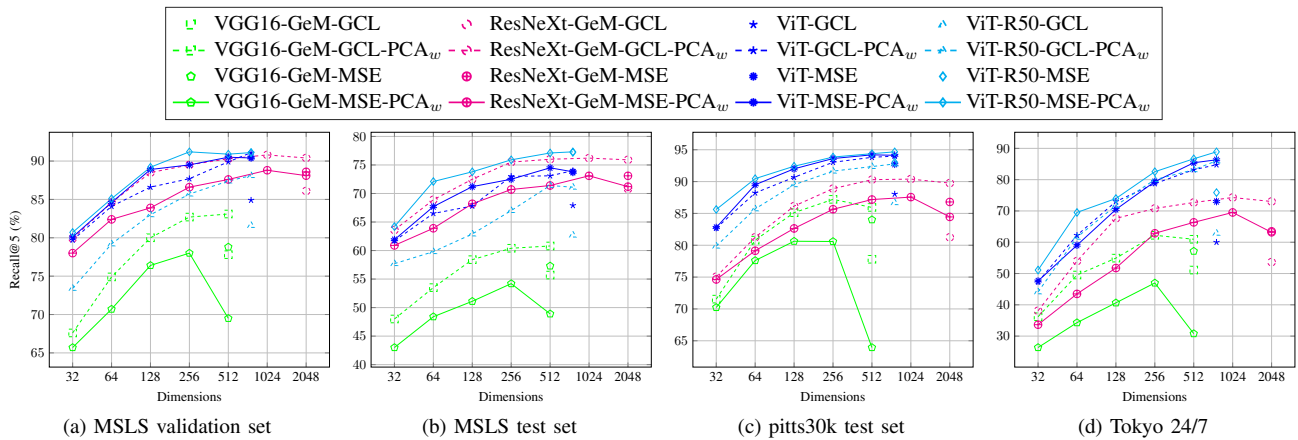

\subsection{Ablation experiments}
\noindent\textbf{Encoder. }
We trained several encoders on the MSLS dataset~\cite{msls} by optimizing the MSE loss, and studied the impact of transformers and convolutional backbones. Following~\cite{Arandjelovic2017,GCL} we chose the following encoders: NetVLAD with a VGG16 backbone, and fully-convolutional backbones VGG16 and ResNeXt with GeM pooling~\cite{radenovic2018fine}. We also considered two transformers, namely a Vision Transformer (ViT) and a Hybrid Vision Transformer (ViT-R50) with knowledge distilled from ResNet50, without a pooling layer nor a projection head. All encoders (and the VGG16 encoder of NetVLAD) were pre-trained on ImageNet. 

In Table~\ref{tab:ablation}, we report results on the MSLS validation and test sets and the generalization performance on the Pitts30k and Tokyo24/7 benchmarks. In all cases, the transformers achieve higher results, with the hybrid ViT-R50 models achieving the highest performance in all considered datasets. However, the ablation results support that the MSE loss can be applied to any architecture for VPR successfully.

\begin{figure}[!t]
    \centering
    \includegraphics[height=2.5cm]{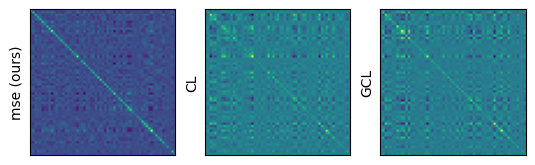}
    \caption{Example of the covariance matrices of features (from the MSLS validation set) learned with MSE and contrastive losses.}
    \label{fig:cov}
    \vspace{-4mm}
\end{figure}

\noindent\textbf{PCA and whitening. }
We study the effect of PCA and whitening on our descriptors~\cite{Arandjelovic2017, radenovic2018fine}. We observed that although the whitening does not lead to significantly better performance on the MSLS dataset, it can help the generalization capabilities of the MSE-trained descriptors, especially in the case of the Tokyo 24/7 datasets (see Table~\ref{tab:ablation}). 
%
We report the results achieved by applying whitening and PCA to decrease the size of the descriptors in Figure~\ref{fig:whitening}. 
For convolutional architectures, PCA whitening leads to a boost in the retrieval performance on the MSLS dataset, with less improvement in the out-of-distribution test benchmarks Pitts30k and Tokyo 24/7. For the MSE-trained models that use VGG16 as backbone (VGG16-GeM and NetVLAD), the PCA and whitening impact negatively on the results, especially in generalization tests. 
The whitening tends to improve the generalization performance of transformer backbones trained with the MSE loss to the Pittsburgh30 and Tokyo24/7 datasets, while not having an impact on in-distribution tests on the MSLS validation and test sets. This is attributable to the fact that MSE-trained models tend to learn features with lower covariance (see Figure~\ref{fig:cov}). 
PCA of the transformer descriptors contributes to maintaining good performance even when down-scaling the feature space size to 128 dimensions.

\section{Conclusions}
We shifted from the contrastive learning paradigm for visual place recognition and showed that this task can be treated as a regression problem at large-scale with high results. We train Vision Transformers to learn global image descriptors whose distance in a latent space is a direct measure of similarity of the images. This straightforward training scheme results in descriptors that have exceptional ranking capabilities. Furthermore, they do not require complex and time-consuming re-ranking to curate the retrieval results. The training process is data-efficient as it requires a few iterations on a limited view of the data, namely a few thousand training pairs. 
Our training scheme dispenses VPR pipelines from the need for large datasets to learn effective descriptors, and can foster training high-performing models in energy-saving settings.
We achieve better performance than other VPR methods and good generalization to out-of-distribution test sets. We achieved results higher than or comparable with methods that use re-ranking strategies or that are trained using triplet loss, hard-pair mining and that are trained for several epochs.

\bibliographystyle{IEEEtran}
\bibliography{references.bib}

\end{document}